\documentclass[conference]{IEEEtran}
\usepackage{amsmath,amsfonts}
\usepackage{algorithm}
\usepackage{array}
\usepackage{textcomp}
\usepackage{stfloats}
\usepackage{url}
\usepackage{verbatim}
\usepackage{graphicx}
\usepackage{cite}
\usepackage{color}
\usepackage{colortbl}
\usepackage{soul}
\usepackage{multirow}
\usepackage{multicol}
\usepackage{makecell}
\usepackage{booktabs}
\usepackage{hyperref}
\usepackage{amssymb}
\usepackage{bbding}
\usepackage{pifont}
\usepackage{wasysym}
\usepackage{subcaption}
\usepackage{bm}
\usepackage{algpseudocode}
\algrenewcommand\algorithmicindent{1.0em}
\usepackage[utf8]{inputenc}
\usepackage{CJK}
\usepackage{threeparttable}
\usepackage[table]{xcolor}
\usepackage{xspace}
\usepackage[numbers,sort&compress]{natbib}
\usepackage[normalem]{ulem}
\usepackage{caption} 

\newcommand{\ti}{\textit}
\newcommand{\tb}{\textbf}

\begin{document}


\title{SanD-Planner: Sample-Efficient Diffusion Planner in B-Spline Space for Robust Local Navigation }



\author{Jincheng Wang, Lingfan Bao, Tong Yang, Diego Martinez Plasencia, Jianhao Jiao$^{*}$, and Dimitrios Kanoulas%
}

\twocolumn[{
    \renewcommand\twocolumn[1][]{#1} 
    \maketitle
    \begin{center}
        \includegraphics[width=0.8\textwidth]{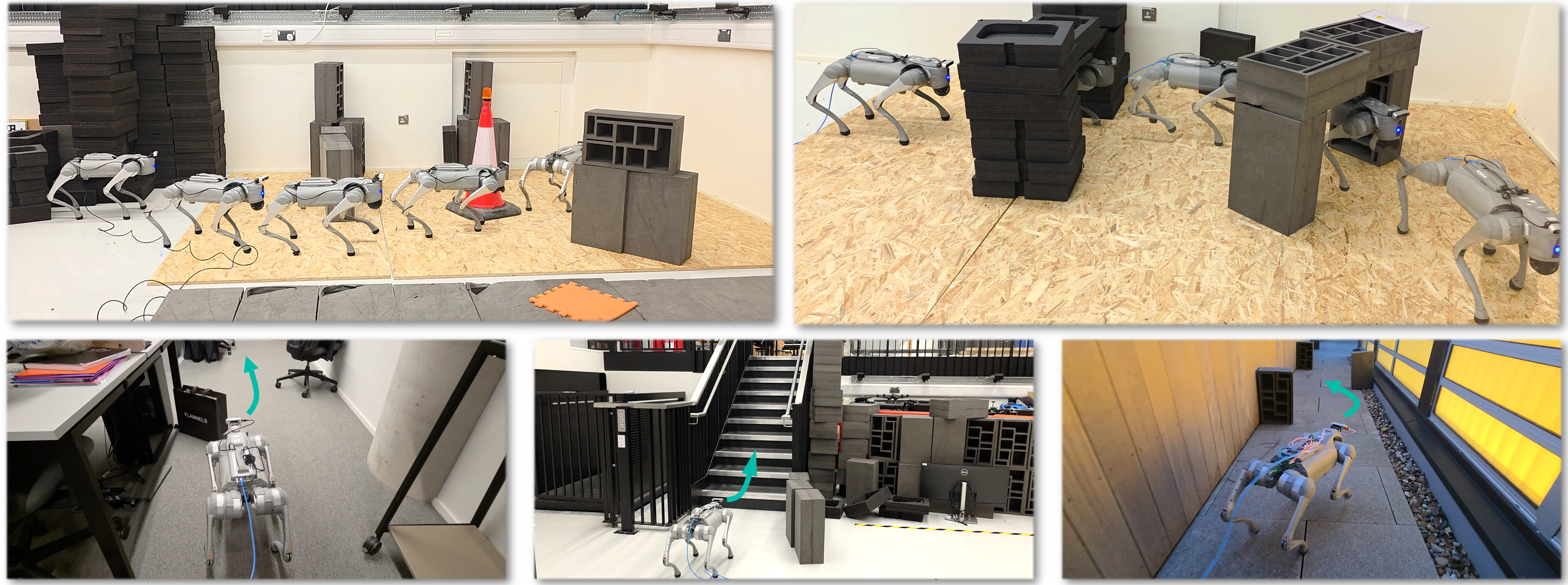}

        \captionof{figure}{Real-world demonstrations of SanD-Planner. Our method enables the robot to traverse cluttered environments with tight clearances, including navigating narrow passages, avoiding obstacles, and climbing stairs.}
        \label{figure1}
    \end{center}

}]

\footnotetext[1]{Corresponding author. {\tt\small jianhao.jiao@ucl.ac.uk} The authors are with the Department of Computer Science, University College London.}

\begin{abstract}
The challenge of generating reliable local plans has long hindered practical applications in highly cluttered and dynamic environments. Key fundamental bottlenecks include acquiring large-scale expert demonstrations across diverse scenes and improving learning efficiency with limited data. This paper proposes SanD-Planner, a sample-efficient diffusion-based local planner that conducts depth image-based imitation learning within the clamped B-spline space. By operating within this compact space, the proposed algorithm inherently yields smooth outputs with bounded prediction errors over local supports, naturally aligning with receding-horizon execution. Integration of an ESDF-based safety checker with explicit clearance and time-to-completion metrics further reduces the training burden associated with value-function learning for feasibility assessment. Experiments show that training with $500$ episodes (merely $0.25\%$ of the demonstration scale used by the baseline), SanD-Planner achieves state-of-the-art performance on the evaluated open benchmark, attaining success rates of $90.1\%$ in simulated cluttered environments and $72.0\%$ in indoor simulations. The performance is further proven by demonstrating zero-shot transferability to realistic experimentation in both 2D and 3D scenes. The dataset and pre-trained models will also be open-sourced.

\end{abstract}

\IEEEpeerreviewmaketitle

\section{Introduction}

The capability for robust navigation in cluttered and dynamic environments remains a fundamental challenge for general-purpose mobile robotics \cite{gul2019comprehensive, yasuda2020autonomous}. Classic navigation frameworks typically utilize a modular pipeline that sequentially executes perception, SLAM, and planning sub-routines \cite{cao2022autonomous, kavraki2002probabilistic, ratliff2009chomp, jiao2026opennavmap}. 
While effective in structured settings, the reliance on precise mapping creates bottlenecks in dynamic environments. Furthermore, the sequential processing induces latency and compounds errors from sensor noise and drift, often compromising the system's ability to safely adapt to abrupt changes.

This has catalyzed a shift toward end-to-end navigation approaches~\cite{shah2023gnm,yang2023iplanner,roth2024viplanner, liu2023vit, liu2024dipper} that map onboard observations directly to actions or trajectories. While deep reinforcement learning (DRL) provides a principled framework for policy acquisition \cite{hoeller2021learning, shi2019end, yadav2023offline, aich2024deep}, it is often hindered by extreme sample inefficiency and the difficulty of reward engineering. 
In contrast, imitation learning (IL) bypasses these issues by learning directly from expert demonstrations. 
Recent advances have shown remarkable results by aggregating large-scale real-world datasets to achieve general navigation behaviors \cite{shah2023vint, sridhar2024nomad,cai2025navdp}. However, the success of these methods remains largely data-driven, relying heavily on the coverage and diversity of demonstrations \cite{yang2023iplanner}. This scaling trend highlights a common pattern: performance improvements often hinge on the expansion of the supervision pipeline rather than on learning effectively from limited data. In specific domains such as marine robotics \cite{peng2025aquaticvision}, this data scarcity is particularly difficult to overcome.

While scaling IL with massive datasets has driven remarkable progress, it often entails substantial computational overhead and longer training cycles. In this paper, we explore a complementary direction by asking whether local navigation can achieve high performance in a ``data desert'' of limited expert demonstrations.
We propose \textbf{SanD-Planner}, a \textbf{Sa}mple-Efficie\textbf{n}t \textbf{D}iffusion-based Local \textbf{Planner} that maps depth observations directly to smooth and parametric paths.
Unlike waypoint-based methods \cite{sridhar2024nomad} that scale the output dimension linearly with the planning horizon, SanD-Planner predicts a fixed set of eight B-spline control points. This representation embeds structural geometric priors as an inductive bias, enabling an extended horizon to mitigate myopic behavior~\cite{schmittle2025long} without increasing learning complexity. 
By construction, B-splines ensure $C^2$ continuity, making smoothness an inherent property rather than a learned feature. Crucially, the local support of B-splines provides robustness against the perceptual uncertainty typical of long-range sensing; noise or occlusions in the far horizon are isolated to distal control points, preventing far-field instabilities from propagating to the immediate execution segment.
Finally, we decouple trajectory generation from safety assessment via an interpretable online critic. By relegating feasibility checks to this explicit geometric module, we reduce learning complexity and allow the policy to focus exclusively on distilling the expert trajectory distribution. The contributions can be summarized as follows:
\begin{itemize} 
  \item \textbf{SanD-Planner}: A sample-efficient, diffusion-based local planner that generates smooth, collision-avoiding paths within a clamped cubic B-spline control-point space. By leveraging this structured representation, it achieves high-performance point-goal navigation using only \textbf{500} expert trajectories, which is \textbf{$\approx 0.25\%$} of the demonstration scale required by current state-of-the-art (SoTA) baselines.

  \item \textbf{Representation Study}: A systematic investigation conducted under a unified training and evaluation protocol, characterizing how trajectory representations, specifically discrete waypoints, interpolating cubic splines, and B-spline control points, impact the performance and sample efficiency of imitation learning-based local planners.

  \item \textbf{Zero-Shot Sim-to-Real Validation}: Successful deployment on a Unitree Go2 quadruped robot across diverse real-world environments, as shown in Fig. \ref{figure1}. This demonstrates robust, depth-only 3D navigation in unseen complex scenes without any real-world fine-tuning. 

  \item \textbf{Reproducibility}: We will release our complete training and evaluation framework, including the dataset and pre-trained models, upon the acceptance.
\end{itemize}

\section{Related Work}
\subsection{Learning-based End-to-End Visual Navigation}
DRL is a prominent category in this domain, employing trial-and-error optimization to eliminate the need for labeled data \cite{hoeller2021learning, shi2019end, yadav2023offline}. However, DRL is notoriously sample-inefficient and relies on complex reward engineering that often proves more difficult than providing expert demonstrations \cite{hussein2017imitation}. IL has emerged as a practical paradigm, enabling policies to learn directly from expert behaviors \cite{bojarski2016end,pfeiffer2017perception,pfeiffer2018reinforced,ehsani2024spoc, shah2023gnm,shah2023vint}. Building on this foundation, recent approaches have explored generative policy (\ti{e.g.,} diffusion policy) learning to model local paths as conditional distributions, capturing the inherent multi-modality of navigation \cite{casado2025navigating,cai2025navdp,gode2025flownav}. 
However, the robustness of IL policies is fundamentally constrained by the diversity and long-tail coverage of training data, especially in complex geometric configurations \cite{roth2024viplanner}. Consequently, pushing the limits of end-to-end navigation has largely depended on scaling expert demonstrations~\cite{schwartz2020green}. This is typically achieved by aggregating large real-world datasets~\citep{shah2023gnm,shah2023vint} or by generating high-quality simulation supervision at scale, rather than learning effectively from limited data. For instance, recent works such as NavDP \cite{cai2025navdp} leverage high-fidelity simulators to generate extensive supervision, utilizing datasets spanning hundreds of kilometers and requiring substantial compute budgets.
In contrast to the trend of data scaling, this work demonstrates that a robust local planning policy can be learned effectively with limited data by injecting structural priors. We leverage B-spline parameterization to explicitly enforce path smoothness and inherent representation-level robustness. By combining this generative policy with an explicit geometric critic for trajectory selection, we decouple safety verification from policy learning. This design enables robust navigation with significantly fewer demonstrations.

\subsection{B-spline Representation in Navigation}
B-splines are standard trajectory representations \cite{zhou2020ego,zhou2019robust,nguyen2021b} in motion planning, where optimizing in a compact control-point space enables efficiency under smoothness and feasibility constraints. Classical optimization-based planners exploit spline structures, such as the convex-hull property, to improve clearance and dynamic feasibility \cite{zhou2020ego,zhou2020robust,ren2025safety}. These results suggest that trajectory parameterization can serve as a powerful inductive bias for learning-based navigation. 
In previous works, many IL policies were supervised using single-step actions \cite{pfeiffer2017perception,pfeiffer2018reinforced,tai2016deep}, making temporal consistency an explicit learning burden \cite{tai2018socially}. To mitigate this, recent planners predict short-horizon waypoint rollouts or action sequences to encourage coherent motion \cite{gode2025flownav,sridhar2024nomad}. Moving beyond discrete sequences, recent approaches predict sparse waypoints and convert them into cubic-spline paths \cite{yang2023iplanner, roth2024viplanner}. In contrast, this work learns local planning directly in a clamped B-spline control-point space, thereby inheriting the continuity and convex-hull properties. Our analysis (Section \ref{b-spline representation}) has demonstrated that B-splines offer distinct advantages for imitation-based planners compared to waypoint or standard cubic-spline parameterizations.

\begin{figure*}[t]
  \centering
  \includegraphics[width=0.95\textwidth]{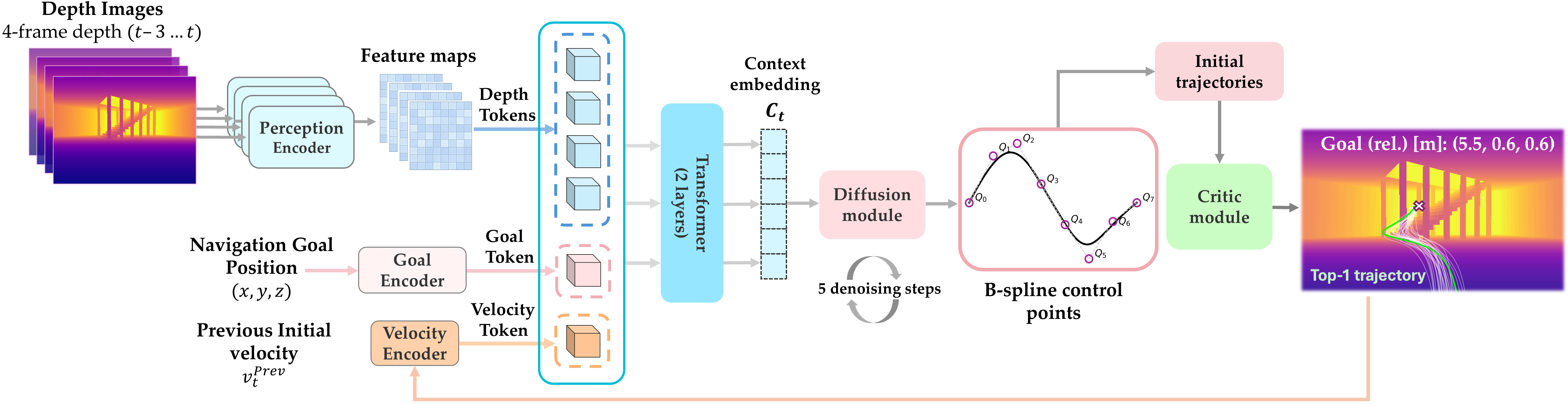}
  \caption{\tb{Overview of SanD-Planner.} The pipeline tokenizes and fuses historical depth images, relative point goals, and previous velocity using a two-layer Transformer encoder. The multi-modal context conditions a diffusion policy to generate B-spline control points through iterative denoising. Finally, an geometric critic module selects the optimal plan from candidates for execution and feeds its initial velocity back to the next planning step to maintain temporal consistency.}
  \label{fig1}
\end{figure*}

\section{Problem Formulation}
This work addresses mapless local navigation in cluttered environments by learning a collision-free goal-reaching policy. At each planning step $t$, the robot receives a short history of four depth images \(
\mathcal{O}_t = \{\mathbf{D}_{t-3}, \mathbf{D}_{t-2}, \mathbf{D}_{t-1}, \mathbf{D}_t\},
\) captured in the robot frame, and receives a three-dimensional relative goal $\mathbf{g}_t = [x, y, z]^{\top}$ expressed in the same frame. To ensure temporal consistency across consecutive replanning cycles, we also incorporate a motion context term $\mathbf{v}_t^{\mathrm{prev}}$ that encodes the heading direction of the plan executed in the previous step. These inputs are projected via domain-specific encoders $f_{(\cdot)}$ and fused into a unified latent context  
$\boldsymbol{\mathcal{C}}_{t} = \Phi\Big(f_{\mathcal{O}}(\mathcal{O}_t),\, f_{g}(\mathbf{g}_t),\, f_{v}(\mathbf{v}_t^{\mathrm{prev}})\Big)$ using an encoder $\Phi$. Instead of predicting waypoint sequences, our planner operates in the structured space of B-spline control points. We parameterize each output local trajectory $\boldsymbol{\tau}_t$ as a sequence of $N$ control points $\mathcal{Q}_t = \{\mathbf{Q}_{t,i}\}_{i=0}^{N-1} \in \mathbb{R}^{N \times 3}$ in the robot frame.

The objective is to generate a smooth, collision-free trajectory $\boldsymbol{\tau}_t$ that efficiently connects the current pose to $\mathbf{g}_t$. To this end, we learn a conditional diffusion model $p_{\theta}(\mathcal{Q}_t \mid \boldsymbol{\mathcal{C}}_{t})$ from expert demonstrations that captures the distribution of feasible control points. During planning, the diffusion policy samples $K$ candidate control-point sets $\{\mathcal{Q}^{(k)}_t\}_{k=1}^{K}$, which generate continuous trajectories $\{\boldsymbol{\tau_t}^{(k)}\}_{k=1}^{K}$. The optimal plan $\boldsymbol{\tau}_t^\star$ is then selected from these candidates by minimizing a task-specific cost $J$, which encompasses safety constraints and efficiency.

\section{Methodology}
As illustrated in Fig.~\ref{fig1}, SanD-Planner presents a two-stage generate-and-select pipeline: 
\ti{1)} trajectory generation within a B-spline control-point space (Sections~\ref{b-spline representation}--\ref{diffusion_policy}), and 
\ti{2)} critic-based selection with explicit geometric objectives (Section~\ref{critic}). 
In the first stage, the policy network uses a conditional diffusion model to sample $K$ candidate control-point sets $\{\mathcal{Q}_t^{(k)}\}_{k=1}^{K}$ conditioned on the context tensor $\boldsymbol{\mathcal{C}}_{t}$. Subsequently, these discrete sets are mapped to continuous local trajectories via B-spline basis functions. In the second stage, a lightweight ESDF-based critic module ranks these candidates by evaluating a task-specific cost function. The trajectory with the lowest cost, $\boldsymbol{\tau}_t^\star$, is selected as the local plan and executed in a receding-horizon manner. The initial heading velocity of the selected trajectory is fed back as \(\mathbf{v}_{t+1}^{\mathrm{prev}}\) to condition the next planning step.

\subsection{Perception and Condition Encoding}
\label{condition embedding}
At each planning step $t$, the agent receives historical observations $\mathcal{O}_t$ to capture environmental dynamics. Each depth frame is encoded using a lightweight ResNet-18 backbone $f_{\mathcal{O}}$ \cite{he2016identity}. To preserve fine-grained geometric details essential for local obstacle avoidance (\ti{e.g.,} obstacle edges or stair steps), we extract spatial feature maps $\mathbf{F} \in \mathbb{R}^{H \times W \times D}$ for each frame from an intermediate layer rather than the final layer.  These spatial features $\mathbf{F}$ are then flattened into a sequence of $H\times W$ visual tokens. To maintain spatial and temporal structure, we augment each token with 2D positional embeddings \cite{liu2021swin} and a learnable temporal embedding to distinguish tokens from different time steps. The resulting visual embedding sequence is denoted as $\mathbf{e}_{\mathcal{O}} = f_{\mathcal{O}}(\mathcal{O}_t) \in \mathbb{R}^{4 \times (HW) \times D}$.

Simultaneously, we project the relative goal $\mathbf{g}_t$ and the previously executed velocity $\mathbf{v}_{t}^{\mathrm{prev}}$ into the same latent dimension $D$ via dedicated Multi-Layer Perceptrons, serving as encoders $f_g$ and $f_v$. This yields the goal token $\mathbf{e}_g = f_g(\mathbf{g}_t)$ and the velocity token $\mathbf{e}_v = f_v(\mathbf{v}_{t}^{\mathrm{prev}})$. If historical velocity is unavailable (\ti{e.g.,} $t=0$), a learnable null token substitutes $\mathbf{e}_v$. All tokens are then concatenated into a sequence and aggregated by a two-layer Transformer encoder $\Phi$. The final output serves as the multi-modal context $\boldsymbol{\mathcal{C}}_{t} = \Phi([\mathbf{e}_{\mathcal{O}}, \mathbf{e}_g, \mathbf{e}_v])$, which is subsequently used to condition the diffusion denoising process.

\subsection{Trajectory Parameterization via B-Spline Control Points}
\label{b-spline representation}
We represent each trajectory as a clamped cubic B-spline which inherently possesses $C^2$ continuity to ensure its smoothness \cite{qin1998general,ding2019efficient,zhou2023tutorial}.
Given a B-spline of degree $p$ and a set of knot vectors $\mathcal{U}=\{{u}_0,\dots,{u}_{N+p}\}$, the trajectory $\tau_t$ is defined as
\begin{equation}
\boldsymbol{\tau}_t(u) \;=\; \sum_{i=0}^{N-1} B_{i,p}(u)\,\mathbf{Q}_{t,i},
\qquad u\in[u_p, u_{N}],
\label{eq:bspline_def}
\end{equation}
where $B_{i,p}(\cdot)$ are the B-spline basis functions. 
We employ a uniform knot vector, anchoring the curve at the endpoints so that $\boldsymbol{\tau}_t({u}_p)=\mathbf{Q}_{t,0}$ and $\boldsymbol{\tau}_t({u}_{N})=\mathbf{Q}_{t, N-1}$.
The control points $\mathcal{Q}_{t}$ are defined in the robot frame, hence $\mathbf{Q}_{t,0}=\mathbf{0}$. 

This parameterization transforms the learning objective from a discrete sequence of waypoints to a compact set of control points. 
To characterize the efficacy of the B-spline representation, we contrast its structured geometric priors against two prevalent IL parameterizations: discrete waypoints \cite{cai2025navdp, gode2025flownav, shah2023gnm} and interpolating cubic splines \cite{yang2023iplanner, roth2024viplanner} in the following content.
As illustrated in Fig.~\ref{fig:bspline_props}, these representations exhibit differences regarding parameter compactness and inherent robustness to perceptual noise. We then discuss three advantages of B-spline parameterization with toy examples:



\begin{figure}[t] 
  \centering
  \begin{subfigure}{1.0\linewidth}
    \centering
    \includegraphics[width=0.95\linewidth]{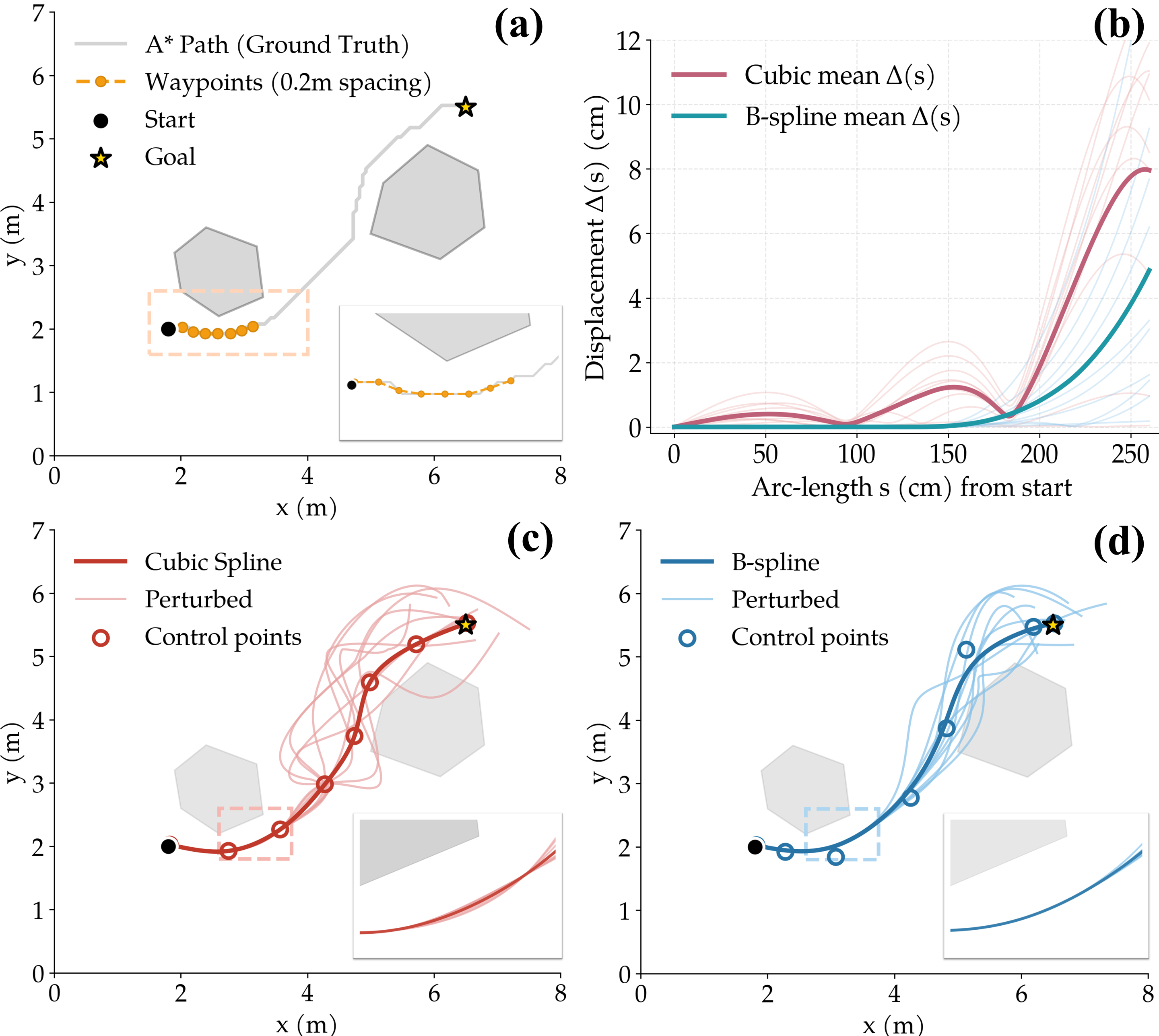} 
  \end{subfigure}
  \caption{Comparison of trajectory representations among discrete waypoints, cubic spline, and B-spline. 
  (a) Ground-truth A* path (gray) and an 8-waypoint sequence (orange, $0.2m$ spacing).
  (b) Mean arc-length displacement $\Delta(s)$ under identical perturbations ($10\times$ disturbances). Over the first $1.5m$, the B-spline deviation stays near zero, whereas the cubic spline oscillates.
  (c, d) We first fit the same ground-truth path with an 8-point interpolating cubic spline (c) and an $8$-point B-spline (d), and then apply the random perturbations to the last four points within a $1m$ radius. Owing to local support and convex hull property, B-splines avoid overshoot and better preserve the global path shape, yielding smaller deviations than cubic splines under the same perturbation range.
  }
  \label{fig:bspline_props}
\end{figure}


\subsubsection{Parameter Efficiency}
As shown in Fig.~\ref{fig:bspline_props}(a) and (c), modeling trajectories with control points instead of dense waypoints yields a compact manifold that simplifies the diffusion process without sacrificing expressivity. Crucially, B-splines facilitate long-horizon planning within a fixed-dimensional output space, effectively avoiding the myopia issues often encountered in traditional methods~\cite{schmittle2025long}.


\subsubsection {Smoothness as Inductive Bias}
Cubic B-splines are $C^2$-continuous, ensuring smooth position, velocity, and acceleration profiles. Critically, this smoothness is a structural property of the representation rather than a feature learned from data. By predicting control points, the policy is constrained to output smooth trajectories even when trained on limited or noisy expert demonstrations, thereby simplifying the learning objective and enhancing trackability for downstream control.

\subsubsection {Representation-Level Robustness} 
Depth sensing often suffers from partial, noisy observations and occlusions, which compromise the reliability of long-range planning. B-splines offer representation-level robustness against this issue through two key properties.

\tb{Local Support:} A B-spline segment depends only on $p+1$ local control points. Consequently, noise-induced deviations in distal control points remain isolated and do not reshape the entire path. This property aligns naturally with receding-horizon planning, allowing the robot to execute a stable near-horizon segment while iteratively correcting far-horizon errors in subsequent steps. In contrast, globally coupled interpolating splines propagate local perturbations across the entire trajectory, potentially destabilizing immediate actions.

\tb{Convex Hull Property:} To ensure safety under distribution shifts, the representation should prevent prediction errors from amplifying into large path deviations. Since B-spline basis functions $B_{i,p}(u)$ satisfy the partition of unity (\textit{i.e.,} $\sum B_{i,p}(u) = 1, B_{i,p} \ge 0$), every point on the trajectory $\boldsymbol{\tau}_t(u)$ lies within the convex hull of its control points. For a predicted set with errors $\mathbf{Q}'_{t,i} = \mathbf{Q}_{t,i} + \Delta\mathbf{Q}_{t,i}$, the trajectory deviation is uniformly bounded by the maximum control-point error:
\begin{equation}
\max_{u} \bigl\|\boldsymbol{\tau}'_t(u) - \boldsymbol{\tau}_t(u)\bigr\| \le \max_i \bigl\|\Delta\mathbf{Q}_{t,i}\bigr\|.
\label{eq:bspline_uniform_bound}
\end{equation}

Eq.~\eqref{eq:bspline_uniform_bound} shows that B-splines prevent error amplification by providing a strict upper bound on path deviation. As shown in Fig.~\ref{fig:bspline_props}, B-splines preserve the global trajectory shape even under perturbation, whereas interpolating splines often suffer from Runge's phenomenon \cite{fornberg2007runge}, where small nodal errors induce disproportionately large overshoots.

\subsection{Diffusion Policy for Local Trajectory Generation}
\label{diffusion_policy}

\subsubsection{Diffusion in B-Spline Control-Point Space}
We model local planning as conditional generation in B-spline control-point space. Given the conditioning context $\boldsymbol{\mathcal{C}}_{t}$, the generator samples a set of control points $\mathcal{Q}_t$, which is deterministically mapped to a continuous trajectory $\boldsymbol{\tau_t}$ via the B-spline representation.
Let $\bar{\mathcal{\mathbf{Q}}}_t\in\mathbb{R}^{N\times 3}$ denote the expert control points obtained by fitting a B-spline to an expert trajectory, and let $\mathbf{x}_0:= \mathrm{vec}(\bar{\mathcal{\mathbf{Q}}}_t)$ be the corresponding clean control-point vector.  We train a conditional DDPM~\cite{ho2020denoising} to model $p_{\theta}(\mathbf{x}_0 \mid \mathcal{C}_t)$. With a predefined noise schedule $\{(\alpha_s,\sigma_s)\}_{s=1}^{S}$ satisfying $\alpha_s^2+\sigma_s^2=1$, the forward process is
\begin{equation}
q(\mathbf{x}_s \mid \mathbf{x}_0) = \mathcal{N}\!\bigl(\alpha_s \mathbf{x}_0,\ \sigma_s^2 \mathbf{I}\bigr),
\qquad s\in\{1,\ldots,S\},
\label{eq:ddpm_forward}
\end{equation}
where $S$ denotes the total number of denoising steps and $s$ is the current diffusion step. We employ the $v$-prediction parameterization~\cite{salimans2022progressive} with the target defined as $\mathbf{v}_s := \alpha_s\boldsymbol{\epsilon} - \sigma_s\mathbf{x}_0$, where $\boldsymbol{\epsilon}\sim\mathcal{N}(\mathbf{0},\mathbf{I})$. A denoiser is trained to predict the diffusion target conditioned on the context embedding and timestep.
Specifically, we use a conditional 1D U-Net over the control-point sequence with cross-attention to $\boldsymbol{\mathcal{C}}_{t}$ and timestep embeddings injected into each block, and predict the $v$-parameterization target $\hat{\mathbf{v}} = \hat{\mathbf{v}}_\theta(\mathbf{x}_s,\boldsymbol{\mathcal{C}}_{t},s)$.
The training objective is the standard $v$-prediction loss~\cite{salimans2022progressive}
\begin{equation}
\mathcal{L}_{\text{diff}}(\theta)
= \mathbb{E}_{\mathbf{x}_0,\boldsymbol{\epsilon},s}
\Bigl[\bigl\| \mathbf{v}_s - \hat{\mathbf{v}}_\theta(\mathbf{x}_s,\mathbf{c}_t,s) \bigr\|_2^2 \Bigr].
\end{equation}

During inference, given ${c}_t$, we initialize the diffusion process from Gaussian noise and run $S$ reverse denoising steps to sample $K$ candidate B-spline control-point sets $\mathcal{Q}_t^{(k)} \sim p_\theta(\mathcal{Q}_t \mid \boldsymbol{\mathcal{C}}_{t})$ for $k\in\{1,\dots,K\}$, which are subsequently mapped to continuous local plans $\{\boldsymbol{\tau}_t^{(k)}\}_{k=1}^K$.

\subsubsection{Temporal Consistency}

Receding-horizon planning under partial observability is often inherently multi-modal, where valid paths exist in distinct topological modes \cite{chi2025diffusion} (\ti{e.g.,} bypassing an obstacle from left or right). If consecutive plans switch between these modes, the executed behavior may oscillate, yielding jittery headings and collisions \cite{torne2025learning}. 
To increase temporal consistency in the generated candidates, we augment the conditioning signal with an explicit previous-plan token $\mathbf{v}_t^{\mathrm{prev}}$ representing the heading direction from the previously selected trajectory $\boldsymbol{\tau}_{t-1}^{\star}$.

With a clamped cubic B-spline, the initial heading direction is proportional to the difference between first and second control points, i.e., $\boldsymbol{\dot{\tau}}(0)\propto (\mathbf{Q}_{t-1,1} - \mathbf{Q}_{t-1,0})$. 
In our robot-centric frame, the first control point is anchored at the origin $\mathbf{Q}_{t-1,0}=\mathbf{0}$. The last heading direction is obtained as
$\mathbf{v}_t^{\mathrm{prev}}=\mathrm{norm}(\mathbf{Q}_{t-1,1}^{\star})$, where $\mathbf{Q}_{t-1,1}^{\star}$ denotes the second control point of $\boldsymbol{{\tau}_{t-1}}^{\star}$. To simplify training and stabilize the learning process, we use privileged information for $\mathbf{v}_t^{\mathrm{prev}}$. During training process, $\mathbf{v}_t^{\mathrm{prev}}$ is computed from the ground-truth expert control points $\bar{\mathcal{\mathbf{Q}}}_{t,2}$ rather than from the previous step. For the first planning step, we use a learned null token to indicate the absence of previous-plan information.

\subsection{Safety-Aware Candidate Selection via the Critic Module}
\label{critic}
While the diffusion policy effectively captures the multi-modal distribution of expert behaviors, the sampled trajectories are not guaranteed to satisfy safety constraints~\cite{luo2024potential,huang2025diffusionseeder}. To decouple complex distribution modeling from strict safety verification, we employ a generate-and-select pipeline. At each planning $t$, the diffusion policy samples $K$ candidate control-point sets $\{\mathcal{Q}^{(k)}_t\}_{k=1}^{K}$ conditioned on the context \(\boldsymbol{\mathcal{C}}_{t}\), which generate continuous trajectories $\{\boldsymbol{\tau_t}^{(k)}\}_{k=1}^{K}$. These are evaluated by an explicit geometric critic, which selects the optimal one $\boldsymbol{{\tau}_t^\star}$ by minimizing a cost function $J$:




\begin{equation}
  \boldsymbol{\tau}_t^\star
  \;=\;
  \underset{\boldsymbol{\tau}\in\{\boldsymbol{\tau}^{(k)}_t\}_{k=1}^{K}}{\arg\min}\;
  J\!\left(\boldsymbol{\tau}\right).
  \label{eq:select_best_tau}
\end{equation}

In contrast to learning-based trajectory ranking~\cite{cai2025navdp} which requires extensive training and often fails under distribution shifts, we propose an explicit, analytic geometric critic. 
This module leverages real-time sensor data for robust safety verification without additional supervision. 
We discretize each candidate spline $\boldsymbol{\tau}^{(k)}_t$ into $M$ waypoints $\{\mathbf{x}_j\}_{j=0}^{M-1}$ equidistantly spaced in arc length. The cost $J$ of each path is then evaluated using the following terms. 




\subsubsection{Discounted Safety Cost}
To score the safety of each trajectory candidate, we build a robot-centric ESDF map $E(\mathbf{x})$ using the current depth image $\mathbf{D}_{t}$. Each voxel stores the signed distance to the nearest obstacle surface, where positive values indicate free space and negative values lie inside obstacles.  
We prioritize obstacle clearance.
Since the planner operates in a receding-horizon manner and executes only the near-horizon segment before replanning, later waypoints should not be weighted equally. We design a temporal discount factor $\gamma \in (0,1)$ to prioritize near-field safety:
\begin{equation}
    J_{\text{esdf}}\bigl(\tau_t^{(k)}\bigr)
    = \frac{1}{\sum_{j=0}^{M-1}\gamma^{j}}
      \sum_{j=0}^{M-1}\gamma^{j}
      \max\bigl(0,\ d_{\text{safe}} - E(\mathbf{x}^{(k)}_j)\bigr),
\end{equation}
where $E(\mathbf{x})$ is the ESDF value at position $\mathbf{x}$, and $d_{\text{safe}}$ is the desired safety margin. This formulation imposes a weighted penalty on clearance violations, placing greater importance on the near-horizon segment of the execution path.


\subsubsection{Path Efficiency}

To encourage the efficiency of the selected plan, we penalize unnecessary detours and prioritize candidates that terminate closest to the local goal $\mathbf{g}_t$.
 \begin{equation}
\begin{aligned}
    J_{\text{len}}\bigl(\boldsymbol{\tau}^{(k)}_t\bigr)
    &= \sum_{j=0}^{M-2} \left\|\mathbf{x}^{(k)}_{j+1} - \mathbf{x}^{(k)}_j\right\|_2, \\
    J_{\text{goal}}\bigl(\boldsymbol{\tau}^{(k)}_t\bigr)
    &= \left\| \mathbf{x}^{(k)}_{M-1} - \mathbf{g}_t \right\|_2.
\end{aligned}
\end{equation}

The total trajectory cost for each candidate path is a weighted sum of the three terms:
\begin{equation}
    J\bigl(\tau_t^{(k)}\bigr)
    = \lambda_{\text{1}}\, J_{\text{safe}}\bigl(\tau_t^{(k)}\bigr)
    + \lambda_{\text{2}}\, J_{\text{len}}\bigl(\tau_t^{(k)}\bigr)
    + \lambda_{\text{3}}\, J_{\text{goal}}\bigl(\tau_t^{(k)}\bigr),
\end{equation}
where $\lambda_{\text{1}}, \lambda_{\text{2}}, \lambda_{\text{3}} > 0$ balance safety and efficiency.




\section{Experiments}
\label{sec:experiments}

This section conduct experiments to answer these questions:

\begin{itemize}
    \item \textbf{Q1.} How well is SanD-Planner compared to other vision-based baselines? (Section \ref{Quantitative Performance Comparison})
    
    \item \textbf{Q2.} How does SanD-Planner scale with the number of expert trajectories (\ti{i.e.,} sample efficiency)? (Section \ref{Sample Efficiency and Data Scaling})
    
    \item \textbf{Q3.} How does trajectory representation affect planning performance? (Section \ref{Ablation study on Trajectory Representation})
    
    \item \textbf{Q4.} How does the velocity token affect planning consistency? (Section \ref{v_token_subsection})
    
    \item \textbf{Q5.} Can SanD-Planner generalize zero-shot to real-world and unseen complex environments? (Section \ref{real_world_section})
    
\end{itemize}

\subsection{Experimental Setup}

\subsubsection{Training Dataset and Implementation Details}

Our training dataset is collected entirely in simulation across 10 diverse environments, including two photorealistic Matterport3D scenes \cite{chang2017matterport3d} and eight Gazebo worlds comprising tunnels and indoor layouts \cite{cao2022autonomous}. To facilitate $3$D local planning, we explicitly incorporate multi-level structures (\ti{e.g.,} stairs) and low-clearance obstacles. We collect $500$ navigation episodes using the PCT-planner \cite{yang2024efficient} as the expert to generate collision-free $3$D trajectories while logging the egocentric depth stream. 
To maximize data efficiency, we construct training pairs by randomly sampling sub-trajectories from each expert episode. Each segment is defined by a random start pose and a future waypoint along the same expert path, which is then fitted with a clamped cubic B-spline using eight control points. 
This resampling strategy generates diverse and informative labels from a limited set of demonstrations, significantly enhancing the policy's sample efficiency.

The model is trained on a desktop equipped with an NVIDIA RTX$4080$ GPU, requiring approximately $5$ hours to converge on the $500$-trajectory dataset. During inference, candidate trajectories are sampled using the deterministic DPM-Solver++ \cite{lu2025dpm}, achieving an operational frequency of approximately $8Hz$. For real-world validation in Section \ref{real_world_exp}, the system is deployed on a laptop with an RTX$4090$ to ensure low-latency planning during autonomous navigation.



\begin{figure}[t]
  \centering
  \begin{subfigure}[t]{0.33\linewidth}
    \centering
    \includegraphics[width=\linewidth]{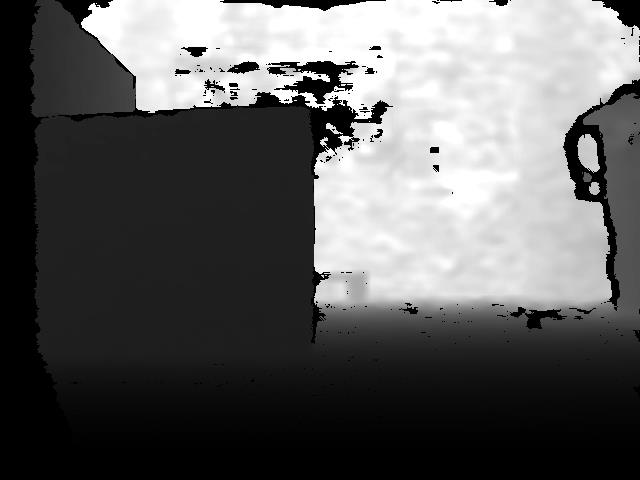}
    \caption{Real}
  \end{subfigure}\hfill
  \begin{subfigure}[t]{0.33\linewidth}
    \centering
    \includegraphics[width=\linewidth]{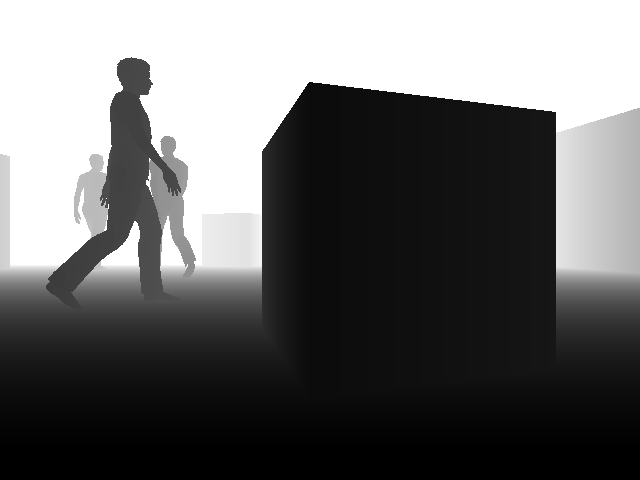}
    \caption{Sim}
  \end{subfigure}\hfill
  \begin{subfigure}[t]{0.33\linewidth}
    \centering
    \includegraphics[width=\linewidth]{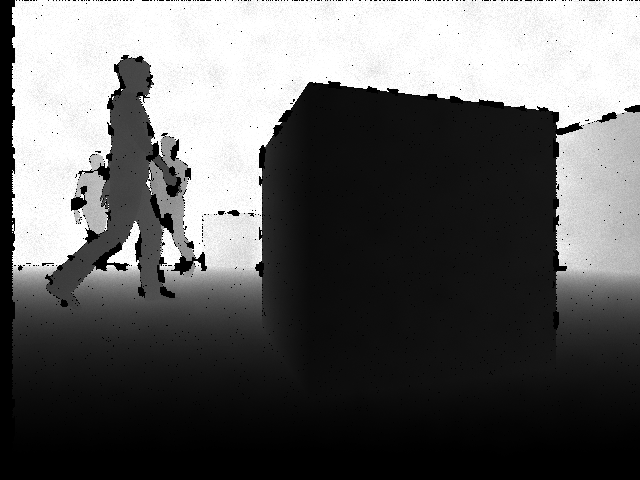}
    \caption{Sim + noise}
  \end{subfigure}
  \caption{Comparison of depth images. (a) Real-world capture. (b) Clean simulated depth. (c) Domain randomization depth.}
  \label{fig:depth_noise}
\end{figure}

\begin{figure}[t]
  \centering
  \includegraphics[width=0.8\linewidth]{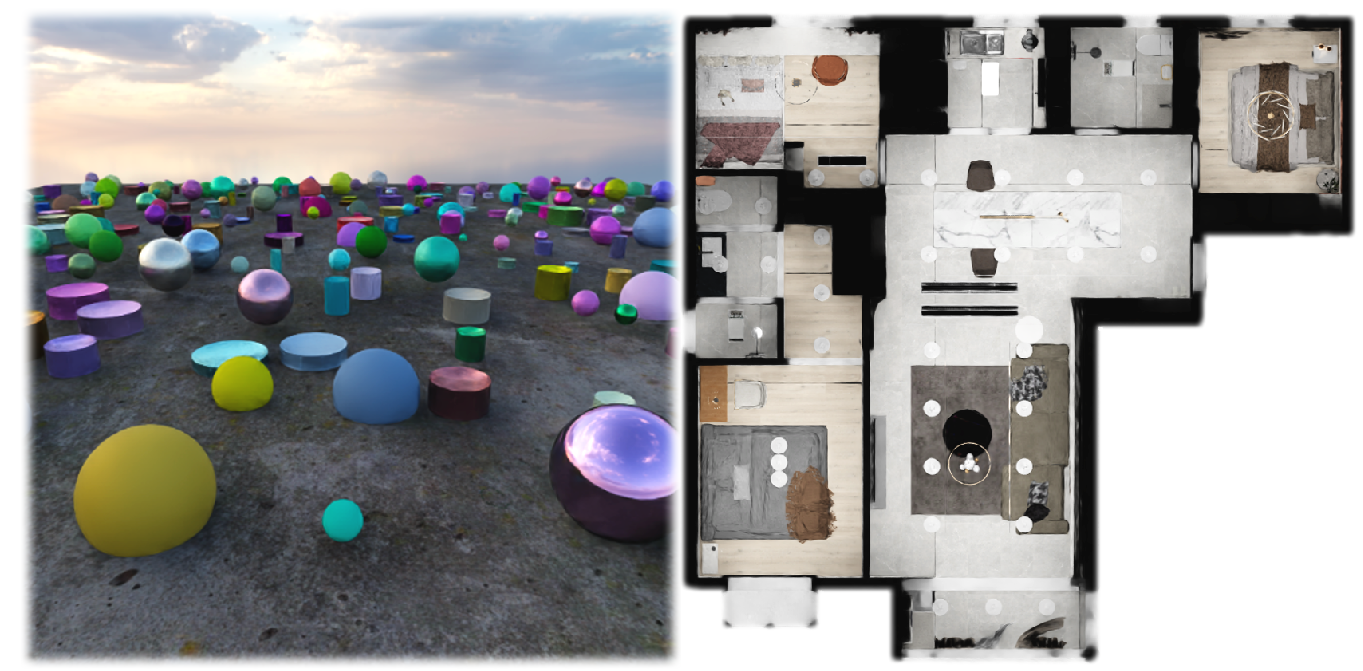}
  \caption{Simulation benchmark environments~\cite{cai2025navdp}. Left: \textbf{ClutteredEnv} provides diverse geometric obstacle layouts. Right: \textbf{InternScenes} features photorealistic indoor settings.}
  \label{benchmark_fig}
\end{figure}

\subsubsection{Observation Domain Randomization}
To bridge the sim-to-real gap, we apply domain randomization to observations during training:
\tb{1) Temporal Latency:} we introduce a temporal freezing mechanism to mimic compute and communication delays. With a probability of $0.1$, we either freeze the observation buffer $\mathcal{O}_t$ to the latest frame or replicate individual frames from their predecessors, simulating signal loss or cold starts.
\tb{2) Sensor Noise:} We emulate noise patterns characteristic of the Intel RealSense D435. Following parameters in \cite{ahn2019analysis}, we inject distance-dependent axial noise that scales quadratically with depth, as visualized in Fig.~\ref{fig:depth_noise}. We also synthesize random pixel dropouts ($0.1$ probability), a stereo-occlusion band on the left margin, and spatially correlated speckle noise. 
Such perturbations force the policy to prioritize structural geometric cues over perfect depth measurements.

\subsubsection{Simulation Benchmark}
We evaluate SanD-Planner using the InternNav Benchmark \cite{cai2025navdp}, which provides photorealistic indoor environments designed to minimize the sim-to-real gap for local obstacle avoidance. The benchmark features diverse assets and predefined start-goal pairs for point-goal navigation. To ensure a fair comparison, all baselines are evaluated under the same protocol as shown in Fig.~\ref{benchmark_fig}.

\begin{table}[t]
    \renewcommand\arraystretch{0.9}
    \renewcommand\tabcolsep{9.7pt}
    \footnotesize
    \centering
    \caption{Quantitative comparison on InternRobotics benchmarks. Results of the baseline methods are taken from \cite{cai2025navdp}.}
    \begin{tabular}{cccc}
        \toprule
        \textbf{Benchmark Environment} & \textbf{Method} & \textbf{SR} $\uparrow$ & \textbf{SPL} $\uparrow$ \\
        \midrule
        \multirow{4}{*}{ClutteredEnv} & iPlanner    & $84.8$ & $83.6$ \\
                                     & ViPlanner    & $72.4$ & $72.3$ \\
                                     & NavDP        & $89.8$ & $\bm{87.7}$ \\
                                     & SanD-Planner & $\bm{90.1}$ & $84.0$ \\
        \midrule
        \multirow{4}{*}{InternScenes} & iPlanner    & $48.8$ & $46.7$ \\
                                     & ViPlanner    & $54.3$ & $52.5$ \\
                                     & NavDP        & $65.7$ & $60.7$ \\
                                     & SanD-Planner & $\bm{72.0}$ & $\bm{63.7}$ \\
        \bottomrule
    \end{tabular}
    \label{local_planner_results}
\end{table}

\begin{table}[t]
  \renewcommand\arraystretch{0.95}
  \renewcommand\tabcolsep{9pt}
  \footnotesize
  \centering
  \caption{Training data and compute resources.}
  \begin{tabular}{cll}
    \toprule
    \textbf{Method} & \textbf{Training Data} & \textbf{Training Time} \\
    \midrule
    iPlanner  & $30$k depth images & $1\times\text{RTX}3090$ Ti for $\approx 20$h \\
    ViPlanner  & $80$k start-goal pairs & $1\times\text{RTX}3090$ for $\approx 6$h \\
    NavDP  & $200$k trajectories & $32\times\text{A}100$ for $24$h \\
    Ours  & $0.5$k trajectories & $1\times\text{RTX}4080$ for $\approx 5$h \\
  \bottomrule
  \end{tabular}
  \label{tab:data_compute}
\end{table}

\subsection{Quantitative Performance Comparison}
\label{Quantitative Performance Comparison}
We evaluate SanD-Planner against three SoTA visual planners: iPlanner \cite{yang2023iplanner}, ViPlanner \cite{roth2024viplanner}, and NavDP \cite{cai2025navdp}, regarding the point-goal navigation task. 
Following the benchmark protocol, we report Success Rate (SR) and Success weighted by Path Length (SPL) \cite{anderson2018evaluation} across $2020$ episodes in ClutteredEnv and $4040$ episodes in InternScenes. 
Notably, despite training on only $500$ trajectories ($\approx\tb{0.25\%}$ of the $200$k trajectories used by NavDP), SanD-Planner achieves competitive or superior performance across both benchmarks (Table~\ref{local_planner_results}).
On the realistic InternScenes, SanD-Planner outperforms NavDP by large margins ($+6.3\%$ SR and $3.0\%$ SPL), demonstrating robust generalization to complex indoor layouts and diverse obstacle distributions. In ClutteredEnv, SanD-Planner achieves the highest $90.1\%$ SR while maintaining a competitive $84.0\%$ SPL. The slight SPL gap compared to NavDP is partly attributable to the B-spline parameterization, which embeds a structural bias toward safer, albeit more conservative, detours. However, analysis in Section~\ref{Sample Efficiency and Data Scaling} suggests this gap arises from the limited data scale rather than methodological constraints and could be further narrowed with additional supervision.

The performance comparison between iPlanner (depth-only) and ViPlanner (depth and semantics) highlights the trade-offs of multi-modal inputs. While ViPlanner leverages semantics in InternScenes ($+5.5\%$ SR over iPlanner), it suffers a severe performance degradation in ClutteredEnv ($-12.4\%$ SR), where primitive obstacles lack well-defined semantic categories. This underscores the potential brittleness of relying on semantic cues under distribution shifts. Crucially, SanD-Planner outperforms ViPlanner on both benchmarks using depth observations alone. These results suggest that depth information provides sufficient geometric cues for general 2D collision avoidance, whereas semantics may degrade model generalization in OOD scenes when training data is limited.

Finally, we distinguish our results from iPlanner and ViPlanner due to their different training paradigms; these methods learn differentiable cost maps from discrete samples (\ti{e.g.,} $30$k images) rather than continuous expert trajectories. 
NavDP, being a full-trajectory IL-based method, serves as the most direct baseline. As detailed in Table~\ref{tab:data_compute}, SanD-Planner demonstrates superior efficiency, requiring orders of magnitude less training data and compute to achieve SoTA performance.

\begin{figure}[t]
    \centering
    \includegraphics[width=1\linewidth]{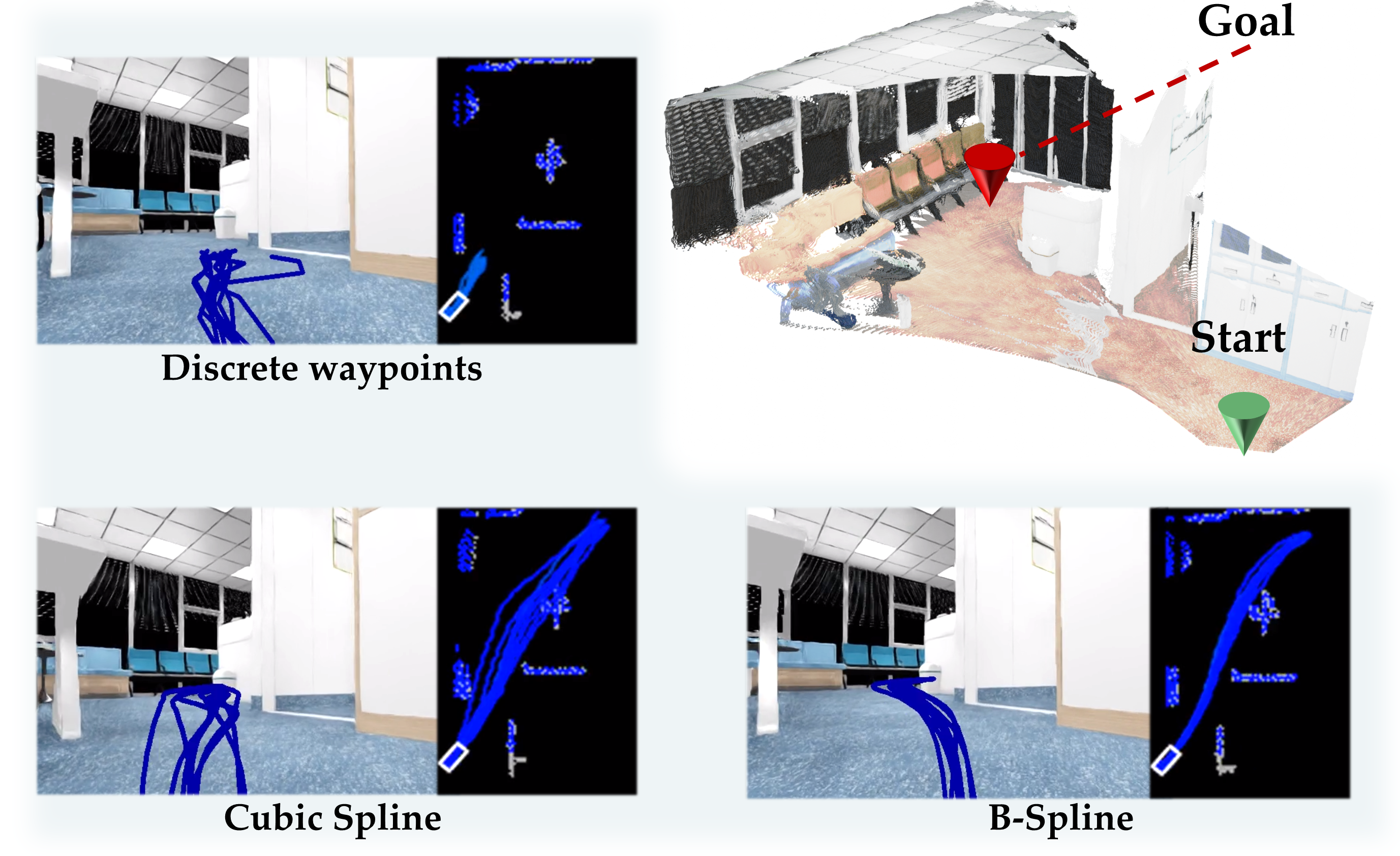}
    \caption{Comparison of trajectory representations in simulation. Top-right: the scene reconstruction with the same start and goal. For each representation (discrete waypoints, cubic spline, and B-spline), we show the ego-view image and a bird-view radar visualization with $10$ candidate trajectories overlaid.}
    \label{fig:repr_qual} 
\end{figure}

\begin{table}[t]
  \renewcommand\arraystretch{0.9}
  \renewcommand\tabcolsep{11.5pt}
  \footnotesize
  \centering
  \caption{Ablation on trajectory representation. Average SR and SPL on two ClutterEnv and two InternScenes scenes.}
  \label{tab:repr_ablation}
  \begin{tabular}{cccc}
    \toprule
    \textbf{Benchmark} & \textbf{Representation} & \textbf{SR} [\%]$\uparrow$ & \textbf{SPL} [\%]$\uparrow$ \\
    \midrule
    \multirow{3}{*}{ClutterEnv}
    & Waypoints         & $75.5$ & $65.8$ \\
    & Cubic spline      & $75.0$ & $70.6$ \\
    & B-spline (Ours) & \boldsymbol{$93.0$} & \boldsymbol{$83.8$} \\
    \midrule
    \multirow{3}{*}{InternScenes}
    & Waypoints         & $75.5$ & $67.8$ \\
    & Cubic spline      & $72.0$ & $65.8$ \\
    & B-spline (Ours) & \boldsymbol{$83.5$} & \boldsymbol{$72.1$} \\
    \bottomrule
  \end{tabular}
\end{table}

\subsection{Sample Efficiency and Data Scaling}
\label{Sample Efficiency and Data Scaling}
\begin{figure}[t]
    \centering
    \includegraphics[width=0.85\linewidth]{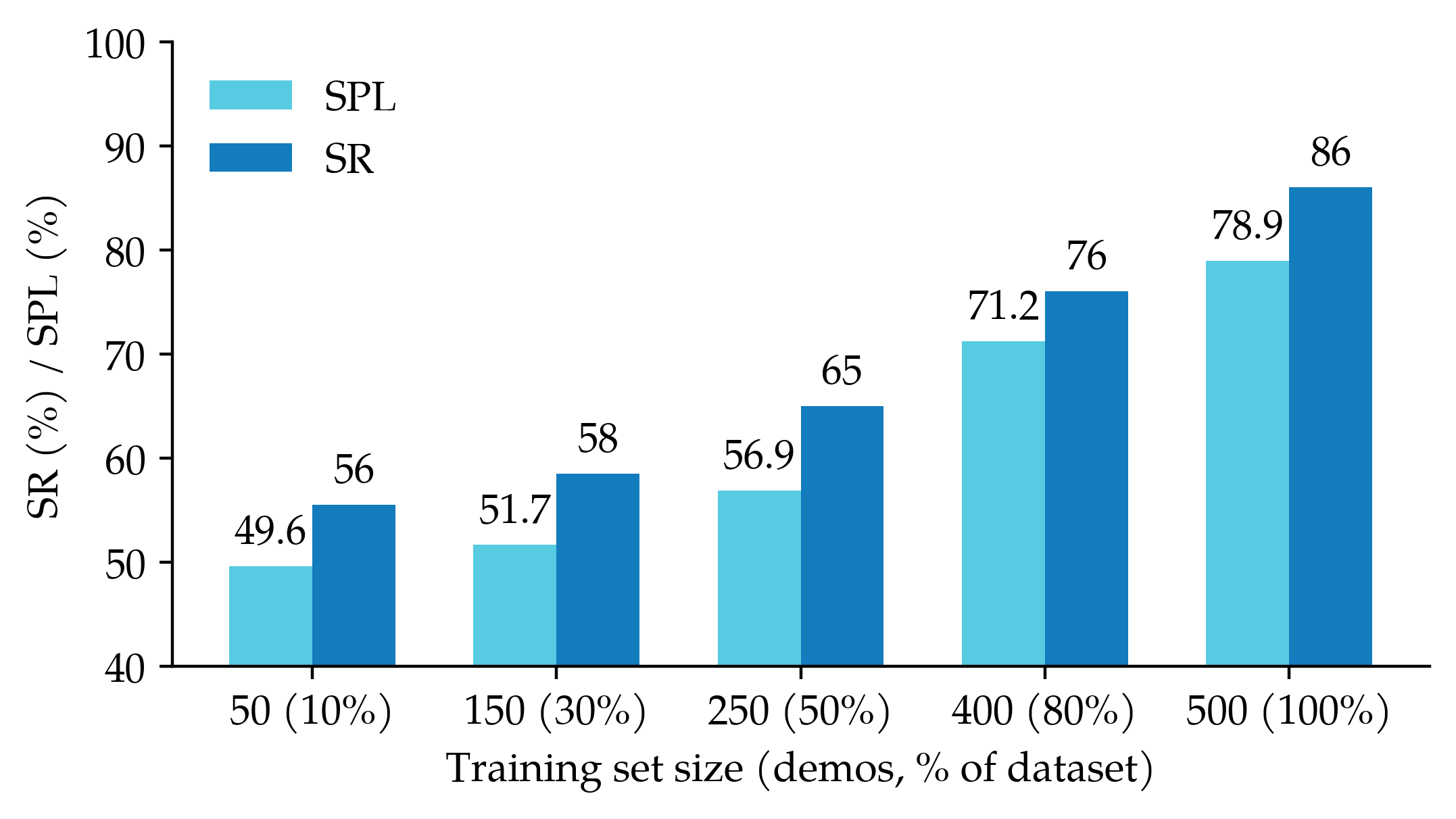}
    \caption{Ablation study on training data size in ClutterEnv. SR and SPL improve consistently with training dataset size.}
    \label{fig:data_scaling}
\end{figure}

We investigate the sample efficiency of SanD-Planner by training on trajectory-level subsets ranging from $10\%$ to $100\%$ of the full dataset. 
As shown in Fig.~\ref{fig:data_scaling}, our approach exhibits monotonic performance gains as the number of demonstrations increases. Notably, the model is highly efficient even in data-scarce regimes, achieving a non-trivial $55.5\%$ SR with only $50$ expert trajectories ($10\%$ data). 
With $250$ trajectories ($50\%$ data), the model attains $65.0\%$ SR, recovering $76\%$ of its peak performance.
While basic collision avoidance is established with minimal data, the SPL metric continues to improve substantially up to the full dataset scale ($49.6\% \rightarrow 78.9\%$). This suggests that while small datasets suffice for learning safe behaviors, larger demonstration scales are essential for the planner to internalize more efficient bypass strategies and minimize unnecessary detours.



\subsection{Ablation Study on Trajectory Representation}
\label{Ablation study on Trajectory Representation}
To validate the impact of trajectory parameterization, we ablate the output representation while keeping the rest of the navigation pipeline fixed. All variants predict eight anchor points to maintain identical output dimensionality. The discrete waypoint baseline utilizes a fixed $0.2m$ spacing, resulting in a $1.4m$ lookahead. In contrast, both interpolating cubic splines \cite{yang2023iplanner} and our B-spline representation parameterize an extended horizon of approximately $6m$.
Tab.~\ref{tab:repr_ablation} shows that the choice of representation influences performance, with B-splines consistently outperforming the alternatives. In ClutteredEnv (episodes $>20m$), discrete waypoints suffer from the myopic behavior \cite{schmittle2025long}; while they match cubic splines with $\approx 75\%$ SR, they exhibit lower efficiency ($65.8\%$ vs. $70.6\%$ SPL) due to suboptimal detours. Spline-based parameterizations mitigate this by enforcing smoothness and extending the planning horizon within the same parameter budget. 

The performance gap of B-splines highlights the critical role of local support. The interpolated cubic splines are globally coupled, causing prediction errors in occluded far-field regions to propagate backward and destabilize the immediate execution segment. Conversely, B-splines possess compact support, which effectively isolates far-field uncertainty and ensures near-field stability. In the confined InternScenesenvironment, the advantage of long-horizon planning is less pronounced due to heavy occlusions. This explains why myopic waypoints slightly outperform cubic splines in such settings. However, B-splines maintain the best overall performance, balancing near-horizon robustness with path efficiency. 
Qualitative results in Fig.~\ref{fig:repr_qual} further illustrate this robustness: waypoints yield non-smooth, less trackable paths, and cubic splines amplify prediction noise across the entire curve. In contrast, B-splines generate consistent candidates with localized variations, demonstrating lower sensitivity to perceptual uncertainty.

\begin{table}[t]
  \renewcommand\arraystretch{0.9}
  \renewcommand\tabcolsep{9.5pt}
  \footnotesize
  \centering
  \caption{Velocity-token ablation on ClutteredEnv scenes.}
  \label{tab:ablation_vtoken}
  \begin{tabular}{ccccc}
    \toprule
    & \multicolumn{2}{c}{\textbf{w/o token}} & \multicolumn{2}{c}{\textbf{w/ token}} \\
    \cmidrule(lr){2-3}\cmidrule(lr){4-5}
    \textbf{Scene} & \textbf{SR} [\%]$\uparrow$ & \textbf{SPL} [\%]$\uparrow$ & \textbf{SR} [\%]$\uparrow$ & \textbf{SPL} [\%]$\uparrow$ \\
    \midrule
    Scene 1 & $79.0$ & $0.73$ & $\bm{87.0}$ & $\bm{0.7}8$ \\
    Scene 2 & $93.0$ & $0.86$ & $\bm{99.0}$ & $\bm{0.8}9$ \\
    \midrule
    \textbf{Average} & $86.0$ & $0.80$ & $\bm{93.0}$ & $\bm{0.84}$ \\
    \bottomrule
  \end{tabular}
\end{table}

\begin{figure*}[t]
  \centering
  \begin{subfigure}[t]{0.73\linewidth}
    \centering
    \includegraphics[width=\linewidth]{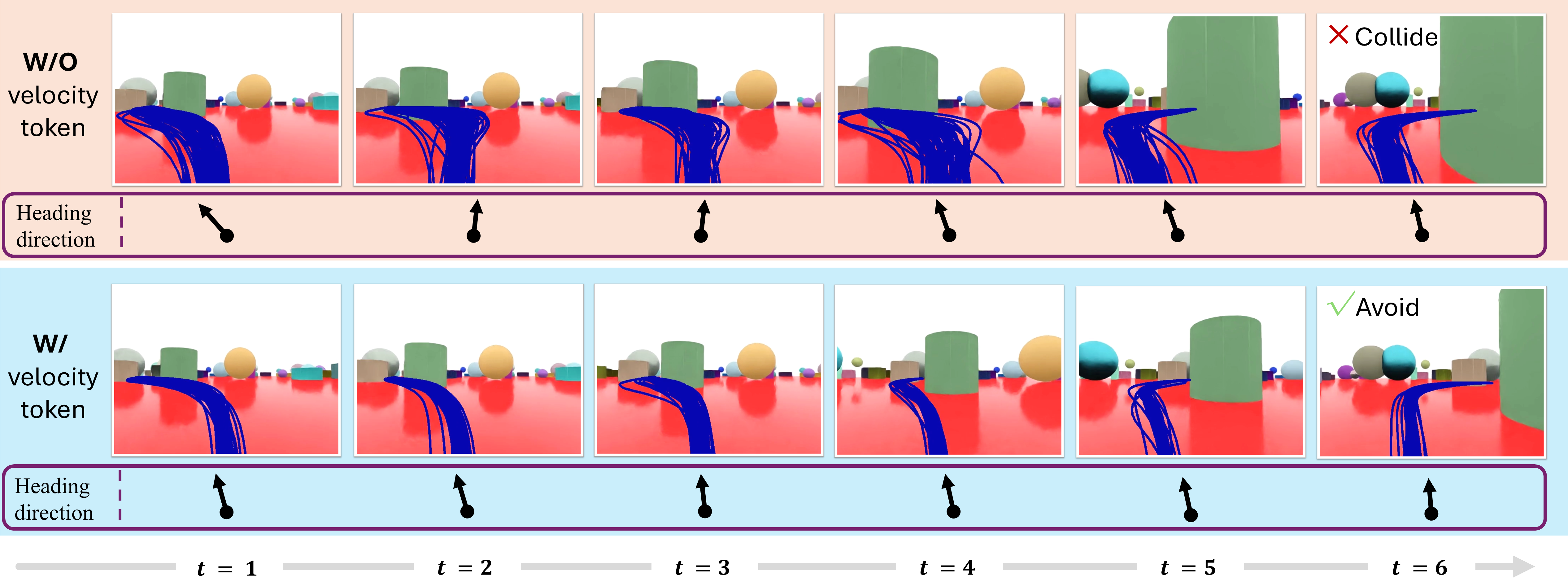}
  \end{subfigure}
  \begin{subfigure}[t]{0.147\linewidth}
    \centering
    \includegraphics[width=\linewidth]{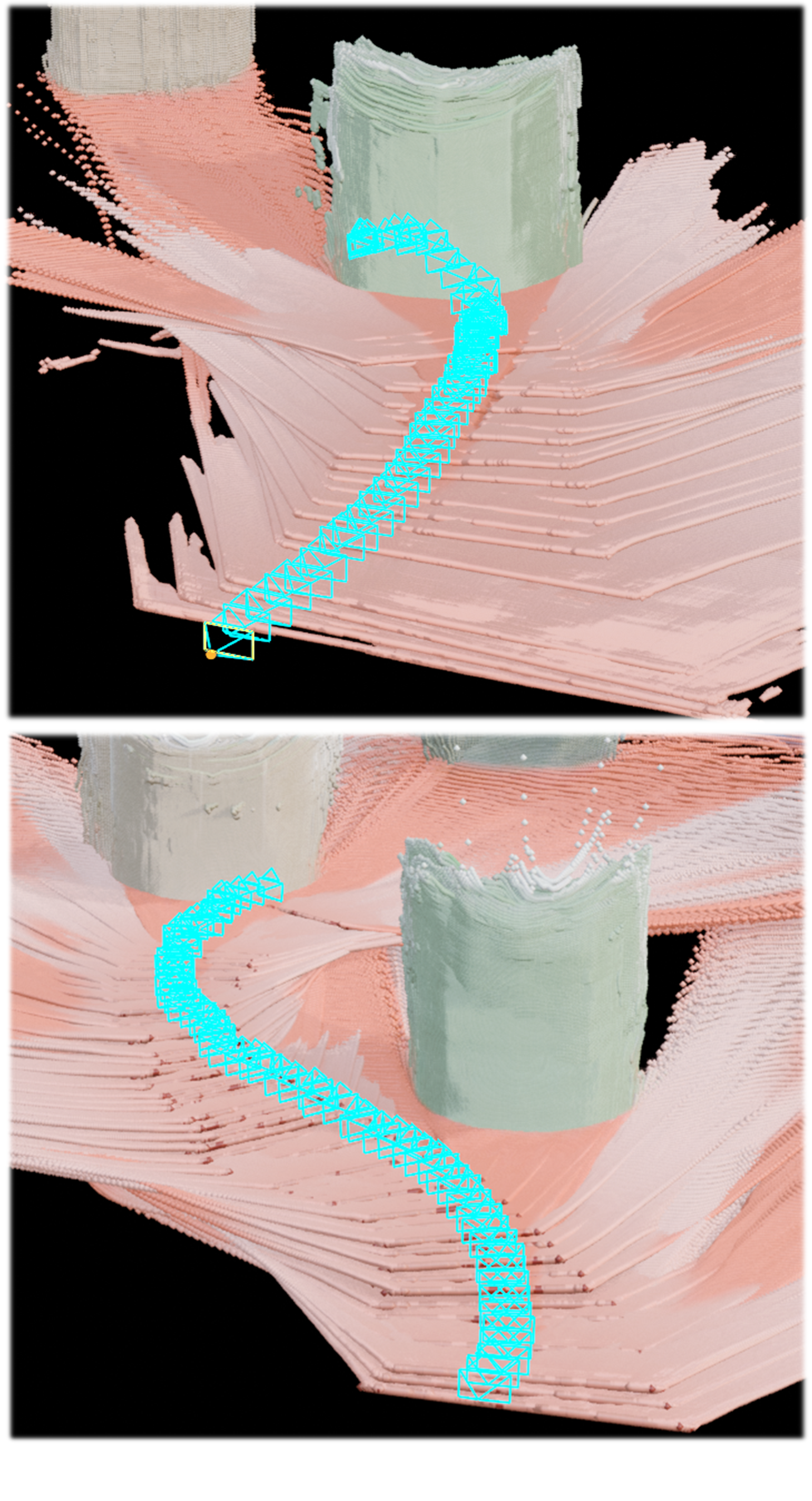}
  \end{subfigure}
  \caption{\textbf{Representative comparison of trajectory consistency with and without the velocity token.} We visualize the diffusion-sampled bypass candidates (blue) at each planning step $t$. Without the velocity token (top), the candidate set remains highly multi-modal and dispersed, and the critic-selected solution alternates between bypass modes across consecutive replans, producing inconsistent heading directions (arrows) and eventually leading to a collision. With the velocity token (bottom), the sampled candidates are more temporally consistent, yielding stable headings and a safe bypass.
  The rightmost panel shows the corresponding 3D reconstructions from Depth Anything 3 \cite{lin2025depth}, where the cyan curve indicates the camera/robot trajectory.}
  \label{fig:vtoken_case}
\end{figure*}

\subsection{Ablation Study on Velocity Token  for Temporal Consistency}
\label{v_token_subsection}
We evaluate the impact of previous plan information on planning consistency by ablating the velocity token $\mathbf{v}_{t}^{\mathrm{prev}}$. 
As reported in Table~\ref{tab:ablation_vtoken}, removing this temporal conditioning leads to a consistent performance drop, which confirms its necessity for trajectory reliability. 
Analysis reveals that collisions without $\mathbf{v}_{t}^{\mathrm{prev}}$ frequently occur near geometrically symmetric obstacles (\ti{e.g.,} pillars), where an example is shown in Fig.~\ref{fig:vtoken_case}.
Under partial observability, such structures induce bimodal distributions of plausible bypass trajectories. Without velocity conditioning, the diffusion model generates dispersed, high-variance candidates at each step. 
Since the critic selects the optimal trajectory independently per step, the planner suffers from topological inconsistency, frequently switching between conflicting bypass modes. This results in oscillatory steering and eventual collisions. 
In contrast, the $\mathbf{v}_{t}^{\mathrm{prev}}$ condition effectively regularizes the generative process, yielding concentrated candidate sets consistent with the robot's established motion and ensuring a stable heading.

\subsection{Real-world Experiments}
\label{real_world_exp}

Real-world experiments are deployed on a Unitree Go2 quadruped robot. The platform features an Intel RealSense D$435$ depth camera with an IR-pass filter, with algorithms executed on a tethered laptop.
We evaluate the system's zero-shot sim-to-real capability without any fine-tuning or domain adaptation. 
The planner operates at approximately $10Hz$, with trajectories tracked by an onboard MPC controller \cite{cai2025navdp}. To facilitate high-frequency replanning in a receding-horizon setting, we employ a warm-start strategy: subsequent planning cycles initialize from the previous solution and undergo partial denoising (starting from step $6$ of $10$). At each step, $K=16$ candidate trajectories are sampled in parallel, with the optimal plan selected by the geometric critic module.

\begin{figure}
    \centering
    \includegraphics[width=0.92\linewidth]{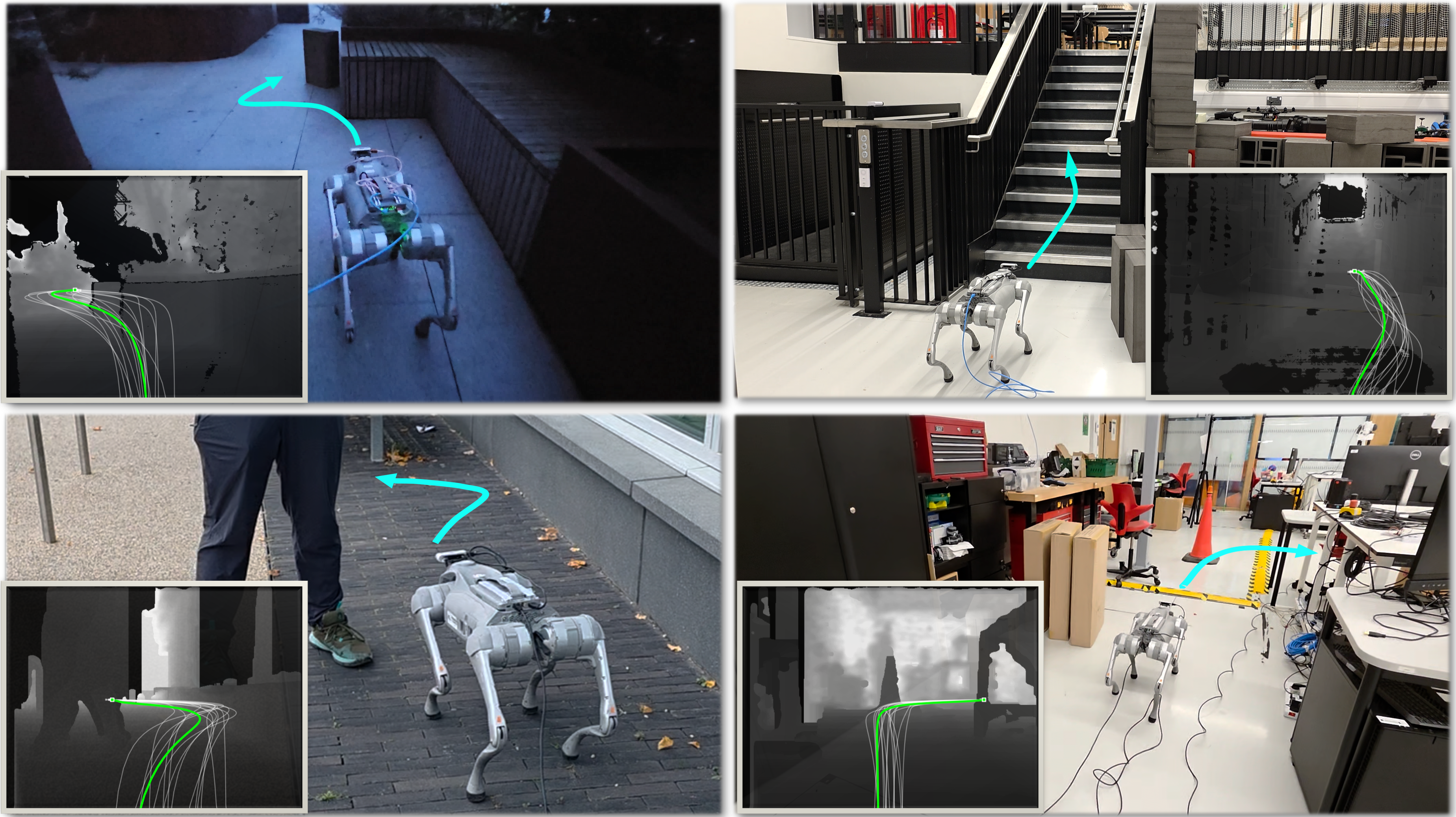}
    \caption{Demonstrations of SanD-Planner in unseen environments. Depth inputs and predicted trajectories are also shown, with the selected trajectory is hightlight in green.}
    \label{real_experiments}
\end{figure}

To assess SanD-Planner's zero-shot generalization, we conduct experiments across diverse scenarios with static and dynamic pedestrians, as shown in Fig.~\ref{figure1} and Fig.~\ref{real_experiments}. 
Static tests include narrow mazes and cluttered office environments with tight clearances. We further evaluate outdoor navigation, where illumination changes and complex geometries present significantly more challenging perception conditions than simulation. Additionally, we demonstrate $3$D navigation capability through stair-climbing trials. Despite these challenges, SanD-Planner generates smooth, collision-avoiding trajectories, confirming the effectiveness and zero-shot capability of our method.

\label{real_world_section}



\section{Conclusion and Future Work}
This paper presents SanD-Planner, a sample-efficient, diffusion-based planner using clamped cubic B-spline parameterization. 
By predicting control points and decoupling the safety check via a critic module, our method reduces learning complexity and data dependency.
Benchmarks show that with only $500$ expert episodes, SanD-Planner matches or surpasses SoTA performance. 
Our findings suggest that: 
1) B-splines isolate sensing uncertainty better than waypoints or interpolating splines, yielding more stable planning; and 2) depth sensing provides sufficient geometric cues for collision avoidance.
Zero-shot real-world deployment further validates our method, enabling $2$D and $3$D navigation tasks such as point-goal navigation and stair traversal without any fine-tuning.
Despite its competitive performance, SanD-Planner is limited by depth sensors regarding small
or specular objects. 
Future work will explore integrating visual foundation models to achieve more robust metric depth estimation and environmental resilience. And even though our primary focus is on sample efficiency, the data scaling experiments reveal the model's untapped potential. We will also plan to leverage this scalability to train on larger, more diverse datasets, aiming to extend the system's generalization across varied environments.

\clearpage
\bibliographystyle{unsrtnat}
\bibliography{references}

\end{document}